\documentclass[11pt,a4paper]{article}
\usepackage{times,latexsym}
\usepackage{url}
\usepackage[T1]{fontenc}

\usepackage[acceptedWithA]{tacl2018v2}

\usepackage{times}
\usepackage{latexsym}
\usepackage{amsfonts}

\usepackage{url}
\usepackage{mathtools}
\usepackage{amsmath}
\usepackage{hyperref}
\usepackage[section]{algorithm}
\usepackage{algpseudocode}
\usepackage{array,multirow}
\usepackage{diagbox}
\usepackage[labelsep=quad,indention=10pt]{caption}
\usepackage[labelfont=bf,list=true]{subcaption}
\usepackage{multirow}

\usepackage{bm}

\def\cD{\mathcal{D}}

\def\WW{\mathbb{W}}

\def\one{\bm{1}}

\usepackage{amssymb}
\usepackage{amsthm}
\newtheorem{definition}{Definition}

 \def\blue#1{}

\usepackage{microtype}

\usepackage{xspace,mfirstuc,tabulary}

\newif\iftaclinstructions
\taclinstructionsfalse
\iftaclinstructions
\newcommand{\instr}
\fi

\iftaclpubformat

\else

\fi

\algnewcommand{\LineComment}[1]{\State \(\triangleright\) #1}

\usepackage{tikz}
\usetikzlibrary{arrows,positioning} 
\tikzset{
    >=stealth',
    punkt/.style={
           circle,
           draw=black,
           text width=1.5em,
           minimum height=1em,
           text centered},
    pil/.style={
           ->,
           shorten <=2pt,
           shorten >=2pt,}
}

\usetikzlibrary{calc}

\title{There Once Was a Really Bad Poet, It Was Automated but You Didn’t Know It}

\author{
Jianyou Wang$^1$, Xiaoxuan Zhang$^1$, Yuren Zhou$^2$, Christopher Suh$^1$, Cynthia Rudin$^{1,2}$ \\
 Duke University \{$^1$Computer Science, $^2$Statistics\} Department\\
 jw542@duke.edu, zhangxiaoxuanaa@gmail.com\\
 yuren.zhou@duke.edu, csuh09@gmail.com, cynthia@cs.duke.edu\\
}

\date{\today}

\begin{document}
\maketitle
\begin{abstract}

Limerick generation exemplifies some of the most difficult challenges faced in poetry generation, as the poems must tell a story in only five lines, with constraints on rhyme, stress, and meter.
To address these challenges, we introduce \textit{LimGen}, a novel and fully automated system for limerick generation that outperforms state-of-the-art neural network-based poetry models, as well as prior rule-based poetry models. \textit{LimGen} consists of three important pieces: the Adaptive Multi-Templated Constraint algorithm that constrains our search to the space of realistic poems, the Multi-Templated Beam Search algorithm which searches efficiently through the space, and the probabilistic Storyline algorithm that provides coherent storylines related to a user-provided prompt word. The resulting limericks satisfy poetic constraints and have thematically coherent storylines, which are sometimes even funny (when we are lucky). 
\end{abstract}

\section{Introduction}


A limerick is a short and catchy 5-line poem that tells a funny, crude or ironic story. It has strict structural constraints such as an AABBA rhyming scheme, a 99669 syllable count, and an anapestic meter pattern \citep{limerickbook}. Writing limericks is a challenging task even for human poets, who have to carefully choose, optimize, and even invent new words to satisfy all of the constraints while incorporating creativity and humor. 

Prior to this paper, there has not been a successful attempt at realistic automatic limerick generation. Perhaps this is because the task is challenging: large-scale neural networks often fail to generate decent limericks because the amount of available human-written limericks to learn from is much smaller than other forms of poetry, and because limericks must follow strict structural, meter, and rhyming constraints. Traditional methods for generating limericks instead hard-code the constraints into a template, so that the constraints are obeyed but the generated poems are all extremely similar (resembling Mad Libs, where one fills words into a single template).

In this paper, we introduce a novel system of algorithms for automatic limerick generation, denoted as \textit{LimGen}. \textit{LimGen} takes a user-specified prompt word and produces a creative and diverse set of limericks related to the prompt. 
Table \ref{tab:eg_expintro} shows some of \textit{LimGen}'s output.
\begin{table}[ht]
\centering\footnotesize
\begin{subtable}{0.49\textwidth}
\centering
\caption{prompt: ``money''}
\begin{tabular}{l}
\hline
There was a greedy man named Todd,\\
Who lost all his money in a fraud. \\
When he returned to work, \\
He was robbed by his clerk, \\
And never could buy a cod.\\
\hline
\end{tabular}
\end{subtable}
\begin{subtable}{0.49\textwidth}
\caption{prompt: ``cunning''}
\centering
\begin{tabular}{l}
\hline
There was a magician named Nick,\\
Who fooled all his family in a trick. \\
When he returned to hide, \\
He was found by his bride, \\
And killed with a magical lipstick.\\
\hline
\end{tabular}
\end{subtable}
\caption{\textit{LimGen} Examples}
\label{tab:eg_expintro}
\end{table}


\textit{LimGen} is a rule-based search method. 
Its main components are: (1) Adaptive Multi-Templated Constraints (AMTC), which constrain \textit{LimGen}'s search to a space of realistic limericks, leveraging knowledge from limerick sentence structures extracted from human poets; (2) the novel Multi-Templated Beam Search (MTBS), which searches the space in a way that fosters diversity in generated poems;
(3) the probabilistic Storyline algorithm, which provides coherent storylines that are thematically related to the prompt word. 

\textit{LimGen} relies on the part-of-speech (POS) limerick templates extracted from a small training set and uses a pre-trained language model to fill words into the templates. We used the 345M version of  pre-trained GPT-2 \citep{radford2019language}, which performs extremely well in unconstrained text generation. However, it is important to note that a language model such as GPT-2, powerful though it may be, is only a plugin module for \textit{LimGen}. Without \textit{LimGen}, GPT-2 alone is completely incapable of generating limericks. 

Through our experiments, we demonstrate that \textit{LimGen} creates a new benchmark for limerick generation, outperforming both traditional rule-based algorithms  
and encoder-decoder style neural networks 
across a variety of metrics, including emotional content, grammar, humor, sensibleness and storyline quality. Furthermore, although \textit{LimGen} is not yet on par with human poets, our experiments show that 43\% of \textit{LimGen}'s output cannot be distinguished from human-written limericks even when directly compared with actual human limericks. 

The main contributions of this paper are the multi-template-guided \textit{LimGen} system and its MTBS search algorithm. Equipped with AMTC, \textit{LimGen} is the first fully-automated limerick generation system that has the ability to write creative and diverse limericks, outperforming existing state-of-the-art methods. Our diversity-fostering beam search (MTBS) is on par with some of the best beam search algorithms in terms of its ability to optimize limerick quality, and it does a significantly better job at fostering diversity than other methods. The code for \textit{LimGen} as well as the complete list of machine-generated limericks used in our experiments are available online \citep{ourcode}. 

From a broader perspective, we have shown that rule-based poetry generation systems that follow a multi-templated approach, as implemented via the Adaptive Multi-Templated Constraints (AMTC) in this work, can perform better than large-scale neural network systems, particularly when the available training data are scarce. Our work indicates that a computational system can exhibit (what appears to be) creativity using domain-specific knowledge learned from limited samples (in our case, POS templates extracted from human-written poems). 
Although we only use templates to capture the part-of-speech structure of limericks, in general, templates can represent any explicit or latent structures that we wish to leverage. Other NLP applications (e.g., biography generation, machine translation, image captioning, machine translation) have also seen revived interest in template-guided approaches \cite{wiseman,yang-2020-soft-template-NMT,POS-DBS,biset}. Thus, it is conceivable that the general framework of \textit{LimGen}, including AMTC and the MTBS algorithm, can be applied to other forms of poetry generation, as well as broader domains in NLP.

\section{Related Literature}\label{sec:informal}

To the best of our knowledge, Poevolve \citep{levy2001computational}, which combines an RNN with an evolutionary algorithm, and the stochastic hill climbing algorithm proposed by \citet{manurung} are the only other serious attempts at limerick generation in the past 20 years. Unfortunately, their implementations did not result in readable limericks, as can be seen in Section \ref{ssec:prior-attempt}.

Traditional rule-based methods in poetry generation are able to enforce hard constraints, such as rhyming dictionaries or part-of-speech (POS) templates \citep{gervas2000, gervas2001, colton2012, ipoet}. \textit{LimGen} is also rule-based, though it has substantially more flexibility and diversity than \citet{colton2012}'s approach which follows a single POS template during poetry generation. Needless to say, the use of adaptive multi-templates makes AMTC the bedrock of \textit{LimGen}. 

Neural language models have recently been able to produce free-style (unconstrained) English poetry with moderate success \citep{hopkins2017automatically, image_poetry}. In Chinese poetry generation \citep{zhang2014chinese, yi2018chinese, wang2016chinese, reinforce}, research has been so successful that it has spurred further efforts in related areas such as sentiment and style-controllable Chinese quatrain generation \citep{mixed-poet, disentangle, sentiment}. However, their large-scale neural network models take advantage of the Chinese quatrain database, which has more than 150k training examples. In contrast, \textit{LimGen} uses less than 300 limericks. Most modern poetry-generation systems are encoder-decoder style recurrent networks (e.g. character-level and word-level LSTMs) with modifications such as various forms of attention mechanisms. \citet{lau-etal-2018-deep} integrated these techniques and proposed  \textit{Deep-speare}, which represents the state-of-the-art for Shakespearean sonnet generation.  In our experiments, we have adapted and re-trained \textit{Deep-speare} for limerick generation. Empirically, it cannot compete with \textit{LimGen}.


For handling rhyming constraints, unlike \citet{ghazvininejad2016generating} and \citet{benhart2018shall} who generate the last word of each line before generating the rest of the line, our proposed Storyline algorithm selects a probability distribution for the last word of each line. 

Beyond poetry generation, templates are often used in other NLP tasks. For biography generation, \citet{wiseman} noted that a template-guided approach is more interpretable and controllable.  \citet{yang-2020-soft-template-NMT} stated that templates are beneficial for guiding text translation. For fostering diversity in generated text, \citet{POS-DBS} found that a part-of-speech template-guided approach is faster and can generate more diverse outputs than the non-templated diverse beam search of \citet{DBS-vijay}. \textit{LimGen}'s Multi-Templated Beam Search (MTBS) generates diverse results by design; it also addresses the problem of degradation of performance when beam size grows larger, which has been a challenge noted in several prior works \citep{degradation-cohen19,vinyals-bs-lesson,koehn-knowles-2017-six}. 


Since all rule-based constraints in \textit{LimGen} are easily enforced by a filtering function, it does not need to borrow any advanced techniques from the area of constrained text generation \citep[e.g.,][]{hokamp-liu-2017-lexically,anderson-etal-2017-guided,post-vilar-2018-fast,ipoet} where constraints are more complicated.

\section{Methodology}\label{sec:method}

We first introduce terminology in Section \ref{ssec:method-term}. We present \textit{LimGen} along with Adaptive Multi-Templated Constraints (AMTC) in Section \ref{ssec:method-AMTC}. We present the MTBS algorithm in Section \ref{ssec:method-mtbs}, and present our \textit{Storyline} algorithm in Section \ref{ssec:method-storyline}. 

\subsection{Terminology}
\label{ssec:method-term}
We first introduce some useful notation for the concepts of (partial) line, (partial) template, language model, filtering function, and scoring function.

\textit{LimGen}'s entire vocabulary is $\WW$ with size $|\WW|$. For word $w\in \WW$, its part-of-speech (POS) is $w.pos$.  The first $t$ words of a complete line $s_i$ forms a \textit{partial line} $s_i^{(t)}$. We store many partial lines $s_i^{(t)}$ with length $t$ in a set $S^{(t)} = \{s_i^{(t)}\}_i $. A new word $w$ concatenated to $s_i^{(t)}$ becomes  $s_{i}^{(t + 1)} = (s_i^{(t)}, w)$. 
A (partial) template is a sequence of POS tags. The (partial) template of line $s$ is $s.pos = (w_1.pos, \dots, w_n.pos)$.  A language model $\mathcal{L}$ processes a (partial) line and gives a probability distribution for the next word. We use $\cD^{(t)}$ to denote the probability distribution at step $t$.  The filtering function $\mathcal{F}$ filters out words that do not satisfy meter and stress constraints by setting the probability mass of these words in  $\cD^{(t)}$ to zero. Since limericks have a specific meter and stress pattern, words that break this pattern are filtered out by $\mathcal{F}$. 

The scoring function for lines is denoted $H(\cdot)$, which is the average negative log likelihood given by the language model. 
Although our search algorithm generally aims to maximize $H(\cdot)$, the language model's scoring mechanism may not be aligned with poetic quality; sometimes a slightly lower scoring poem has better poetic qualities than a higher scoring one. Thus we may find a better poem by sifting through \textit{LimGen's} output, rather than choosing the highest scoring poem.  
\begin{figure*}[t]
    \centering
    \includegraphics[width=\textwidth]{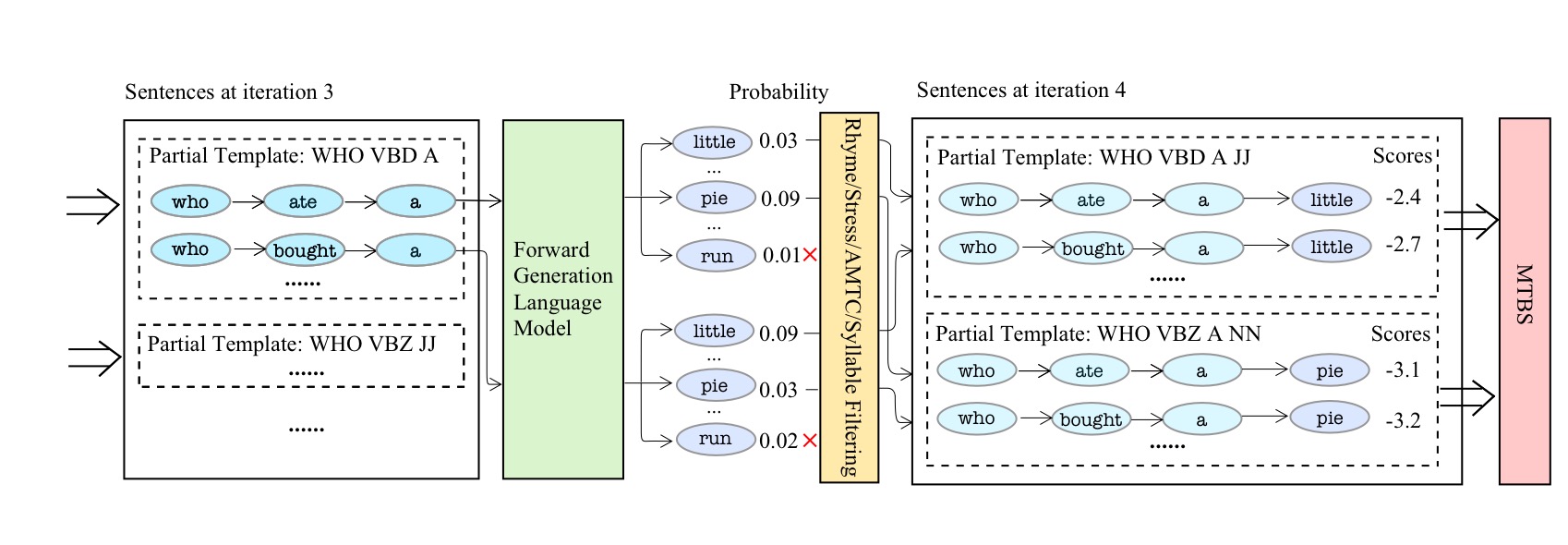}
    \caption{General Framework of \textit{LimGen}}
    \label{fig:LimGen}
\end{figure*}

\subsection{Adaptive Multi-Templated Constraint (AMTC)}
\label{ssec:method-AMTC}
Because the number of limericks that exist in available databases is so limited, we cannot expect that a neural network would learn the POS constraints for a valid limerick. Instead, we use rule-based POS constraints, which are useful in that they ensure the output adheres to the known structure of poetry. The use of adaptive multi-templates makes the poems more diverse and interesting by providing \textit{LimGen} with greater flexibility in pursuing many different templates.  


It may seem natural to choose multiple templates and have the limerick generation process follow each one of them in parallel, but this is inefficient; instead, we start generating from one template, keeping also the set of templates that agree with what we have generated so far. This way, we generate each line by combining a set of related templates. 
Specifically, 
AMTC constrains \textit{LimGen} to consider word $w$ for partial line $s^{(t)}$ only if the template of $s^{(t + 1)} = (s^{(t)}, w)$ matches with a human-written template up to the first $t+1$ tokens. Therefore, the more templates we extract from real poems, the higher the degree of freedom we offer \textit{LimGen}. 

\label{ssec:method-example}
\begin{algorithm}[ht]\small
\caption{\textit{LimGen} with AMTC}
\label{algo:LimGen}
\begin{algorithmic}
\LineComment{\textcolor{blue}{{\scriptsize{ In Section \ref{ssec:method-storyline} the Storyline Algorithm describes how storylines are integrated with \textit{LimGen} to generate last words of each line.}}}}
\State Initialize $S^{(0)} \leftarrow \{[~]\}$;
\For{$t = 0, 1, 2, \ldots$}
\LineComment{\textcolor{blue}{{{\scriptsize{$\tilde{S}^{(t + 1)}$ will store all candidate partial lines of length $t+1$}}}}}
\LineComment{\textcolor{blue}{{{\scriptsize{$S^{(t + 1)}$ will store the chosen partial lines by MTBS}}}}}
\State Initialize $S^{(t + 1)} \leftarrow \varnothing, \tilde{S}^{(t + 1)} \leftarrow \varnothing$;
\For{$s_i^{(t)} \in S^{(t)}$}
\LineComment{\textcolor{blue}{{{\scriptsize{Filter the distribution $\mathcal{L}(s_i^{(t)})$ given by GPT-2}}}}}
\LineComment{\textcolor{blue}{{{\scriptsize{for meter and stress match}}}}}
\State $\cD_{i}^{(t + 1)} \leftarrow  \mathcal{F}( \mathcal{L}(s_i^{(t)}))$;
\For{$w_k \in \WW$ with $\cD_{i}^{(t + 1)}(w_k) > 0$}
\LineComment{\textcolor{blue}{{\text{\scriptsize{AMTC ensures partial template is always the}}}}}
\LineComment{\textcolor{blue}{{\text{\scriptsize{prefix of a viable human template}}}}}
\State \textbf{If} $[s_i^{(t)}, w_k]$ satisfies AMTC:
\LineComment{\textcolor{blue}{{{\scriptsize{if $w_k$ is the last word (i.e., $[s_i^{(t)}, w_k]$'s template}}}}}
\LineComment{\textcolor{blue}{{{\scriptsize{matches a human template), where Storyline }}}}}
\LineComment{\textcolor{blue}{{{\scriptsize{Algorithm contributes to generation of $w_k$}}}}}

\State \hspace{10pt} Concat $[s_i^{(t)}, w_k]$, union with $\tilde{S}^{(t + 1)}$;
\EndFor
\EndFor

\LineComment{\textcolor{blue}{{{\scriptsize{Find top N lines using multi-templated beam search}}}}}

\State ${\tilde{s}^{(t+1)}_1, \dots, \tilde{s}^{t+1}_N}\leftarrow \textit{MTBS}(\tilde{S}^{(t + 1)})$;
\State Union with $S^{(t + 1)}$;
\EndFor\\
\textbf{Output} $S^{t+1}$
\end{algorithmic}
\end{algorithm}
We present the entire \textit{LimGen} system with AMTC in Algorithm \ref{algo:LimGen}. 

We illustrate \textit{LimGen} with the example in Figure \ref{fig:LimGen}.
At the 3\textsuperscript{rd} step of generating the second line, set $S^{(3)}$ contains partial lines $s_1^{(3)} = $\textit{``who ate a''} and $s_2^{(3)} = $\textit{``who bought a''}, which share the same partial template \textit{``WHO VBD A''}.
The probability distributions for the fourth words are
$\cD_1^{(4)} = \mathcal{L}(s_1^{(3)})$ and $\cD_2^{(4)} = \mathcal{L}(s_2^{(3)})$.
One can see how \textit{LimGen} with AMTC does not follow a single template using the example in Figure \ref{fig:LimGen} since the partial template 
\textit{``WHO VBD A''} branches into two distinct partial templates \textit{``WHO VBD A JJ''} and \textit{``WHO VBD A NN''}.

After $\mathcal{F}$ filters out all unsatisfactory words that break the syllable or stress pattern, we obtain two filtered distributions $\tilde{\cD_1}^{(4)} = \mathcal{F}(\cD_1^{(4)})$ and $\tilde{\cD_2}^{(4)} = \mathcal{F}(\cD_2^{(4)})$. We then filter out words that do not satisfy the AMTC. The concatenation step in \textit{LimGen} saves all possible partial lines into a temporary set $\tilde{S}^{(4)}$.

The MTBS algorithm then finds $N$ diverse and high-scoring candidate lines from  $\tilde{S}^{(4)}$ and saves them into  ${S}^{(4)}$. In the next section, we present the MTBS algorithm in detail.

\subsection{Multi-Templated Beam Search (MTBS)}

\label{ssec:method-mtbs}
At iteration $t$,  suppose we have a set of partial lines $S^{(t)}$ with size $N$. Let $\tilde{S}^{(t+1)}$ be the set of all possible one-word extensions of these partial lines. Given the scoring function $H(\cdot)$, a standard beam search would sort $\tilde{S}^{(t+1)}$ in descending order and keep the top N elements. In limerick generation using standard beam search, we also observed the phenomenon documented by \citet{Jurafsky} that most of the completed lines come from a single highly-valued partial line. As mentioned before, the innovation of MTBS over previous diverse beam search papers \citep{Jurafsky,DBS-vijay} is that it calculates a diversity score between (partial) templates (runtime $O(N^2)$), which is more computationally efficient than an approach that assigns a notion of diversity between individual lines (runtime $O(N|\WW|)$, $N\ll|\WW|$, where $N$ is the total number of templates and $|\WW|$ is the vocabulary size). Our proposed diversity score also more accurately captures the diversity between generated lines. The intuition is that if two generated lines have very different templates, they are usually fundamentally different in terms of the progression of the story.

We use a weighted hamming distance to measure the difference between (partial) templates of the same length, denoted as ``diversity score.'' Before formally defining diversity score, we calculate the weights of each POS category. For each POS category, we take the inverse of its percentage of occurrence within all those $n$\textsuperscript{th} line templates (e.g., second line templates) extracted from the $n$\textsuperscript{th} line of one of our human-written limericks from the database. (The POS weights are only calculated once, before we start generating the $n$\textsuperscript{th} line.) We then use the softmax to transform them into weights for each POS, which measure how rare these POS categories are. The softmax nonlinear transformation softly clips the large weights of outlier POS categories that appear only once or twice. More formally we have:
\begin{definition}[\textbf{Part-of-Speech Weight}]
Let $P$ be the set of all POS categories that occur in all the $n$\textsuperscript{th} line complete-line templates extracted from the limerick database. $|P|$ is its size. For $p_i \in P$, the proportion of $p_i$ is $q_i=\frac{\# p_i\textrm{ occurrences}}{\sum_{p_j\in P}\# p_j\textrm{ occurrences}}$, and the weights of $\{p_i\}_{1\leq i \leq |P|}$ are defined as 
$$\{w(p_i)\}_{1\leq i \leq |P|}=\textrm{softmax}\Big(\big\{1/q_i\big\}_{1\leq i \leq |P|}\Big).$$
\end{definition}

\begin{definition}[\textbf{Diversity Score}]
For (partial) templates $T_1=\{pos_{11},\dots, pos_{1n}\}$ and $T_2=\{pos_{21},\dots,pos_{2n}\}$, assume index set $A=\{i | pos_{1i} \neq pos_{2i}\}$, then we define the diversity score (weighted hamming distance) between $T_1$ and $T_2$ as $$\|T_1-T_2\|_{div}=\sum_{i \in A} \max(w(pos_{1i}),w(pos_{2i})).$$
\end{definition}

Consider a scenario where (partial) templates $T_1$ and $T_2$ have different POS categories at index $i$ but both categories are fairly common (for instance, one noun and one verb), and where (partial) templates $T_1$ and $T_3$ also have different POS categories at index $j$ but one or both are rare. Our proposed diversity score will ensure that the diversity between $T_1$ and $T_3$ is greater than the diversity between $T_1$ and $T_2$, which aligns with our intuition. 

In short, given the set of partial lines $\tilde{S}^{(t+1)}$, MTBS will choose $N$ lines, denoted as ${S}^{(t+1)}$, such that they are high-scoring and generated using many diverse templates. Specifically, we divide $\tilde{S}^{(t+1)}$ into $m$ subsets $\{T_1:\tilde{S}_1^{(t+1)},\dots,T_m:\tilde{S}_m^{(t+1)}\}$, where each subset corresponds to a unique (partial) template of length $t+1$. According to scoring function $H(\cdot)$, for each of these subsets $\tilde{S}_i^{(t+1)}$, we calculate its aggregate score $h_i$ by averaging its $n$ highest-scoring lines. For ease of notation, we let $B=\{T_1:h_1,\dots,T_i: h_i, \dots T_m:h_m\}$, and we initialize $A=\emptyset$ to be the set of previously chosen templates. At this point, we shall iteratively determine the order by which lines from these $m$ subsets will be included into ${S}^{(t+1)}$.

\begin{algorithm}[ht]\small
\caption{Multi-Templated Beam Search (MTBS)}
\label{algo:MTBS}
\begin{algorithmic}
\LineComment{\textcolor{blue}{{\scriptsize{The following describes MTBS at iteration $t$}}}}
\State Input $\tilde{S}^{(t+1)}$
\State Initialize $S^{(t+1)} \leftarrow \emptyset$,  $A \leftarrow \emptyset$ ;
\LineComment{\textcolor{blue}{{\scriptsize{$A$ will hold the templates we have chosen}}}}
\State Split $\tilde{S}^{(t+1)}$ by templates into $m$ subsets: 
\State $\{T_1:\tilde{S}_1^{(t+1)},\dots,T_m:\tilde{S}_m^{(t+1)}\}$,
\LineComment{\textcolor{blue}{{\scriptsize{Lines in the same subset share the same template}}}}
\State For each $\tilde{S}_i^{(t+1)}$, we calculate score $h_i$ by averaging its top $n$ lines according to $H(\cdot)$
\State $ B \leftarrow \{T_1:h_1,\dots,T_i: h_i, \dots T_m:h_m\}$;
\LineComment{\textcolor{blue}{{\scriptsize{In $B$, each template corresponds to an aggregate score $h$}}}}
\State Assume $h_j=\max B$, append top $n$ lines according to $H(\cdot)$ from $\tilde{S}_j^{(t+1)}$ to  $S^{(t+1)}$;
\State Delete $h_j$ from $B$, append $T_j$ into $A$;
\While {$|S^{(t+1)}|\leq N-n$ and $B \neq \emptyset$}
\State $x\in \textrm{argmax}_i \Big(h_i \sum_{T_k\in A} \|T_i-T_k\|_{div} \}\Big)$;
\State Append top $n$ lines from $\tilde{S}_x^{(t+1)}$ to  $S^{(t+1)} $;
\State Delete $h_x$ from B, append $T_x$ into $A$; 
\EndWhile
\State Return  $S^{(t+1)} $;
\end{algorithmic}
\end{algorithm}

\begin{figure*}[ht]
\centering
\begin{tikzpicture}[node distance=1cm, auto,]
\node[punkt, fill = blue!30] (y0) {$y_0$};
\node[punkt, above = of y0, fill = orange!30] (S1) {$l_1$}
edge[dotted, thick, <-] (y0);
\node[punkt, right = of y0, fill = green!30] (y2) {$y_2$}
edge[pil, <-] (y0);
\node[punkt, above = of y2, fill = orange!30] (L2) {$l_{1:2}'$}
edge[dotted, thick, <-] (S1)
edge[dotted, thick, <-] (y0)
edge[dotted, thick, ->] (y2);
\node[punkt, right = of y2, fill = green!30] (y3) {$y_3$}
edge[pil, <-, bend left = 30] (y0);
\node[punkt, above = of y3, fill = orange!30] (L3) {$l_{1:3}'$}
edge[dotted, thick, <-] (y2)
edge[dotted, thick, <-] (L2)
edge[dotted, thick, ->] (y3);
\node[punkt, right = of y3, fill = green!30] (y4) {$y_4$}
edge[pil, <-, bend left = 37.5] (y0)
edge[pil, <-, bend left = 30] (y2)
edge[pil, <-] (y3);
\node[punkt, above = of y4, fill = orange!30] (L4) {$l_{1:4}'$}
edge[dotted, thick, <-] (y3)
edge[dotted, thick, <-] (L3)
edge[dotted, thick, ->] (y4);
\node[punkt, right = of y4, fill = green!30] (y5) {$y_5$}
edge[pil, <-, bend left = 45] (y0)
edge[pil, <-, bend left = 37.5] (y2)
edge[pil, <-, bend left = 30] (y3);
\node[punkt, above = of y5, fill = orange!30] (L5) {$l_{1:5}'$}
edge[dotted, thick, <-] (y4)
edge[dotted, thick, <-] (L4)
edge[dotted, thick, ->] (y5);
\node[punkt, right = of y5, fill = green!30] (y1) {$y_1$}
edge[pil, <-] (y5);
\node[punkt, right = of L5, fill = red!30] (L) {$L$}
edge[dotted, thick, <-] (L5)
edge[dotted, thick, <-] (y5)
edge[dotted, thick, <-] (y1);
\end{tikzpicture}
\caption{Directed acyclic graph for generating a limerick $L$ with Storyline algorithm given prompt $y_0$, with solid arrows representing dependency through Storyline distribution \eqref{eq:prob_dist}, shaded arrows representing the generation process of \textit{LimGen}, and the total beam size of MTBS set to be 1 for simplicity. }
\label{fig:storyline_graph}
\end{figure*}
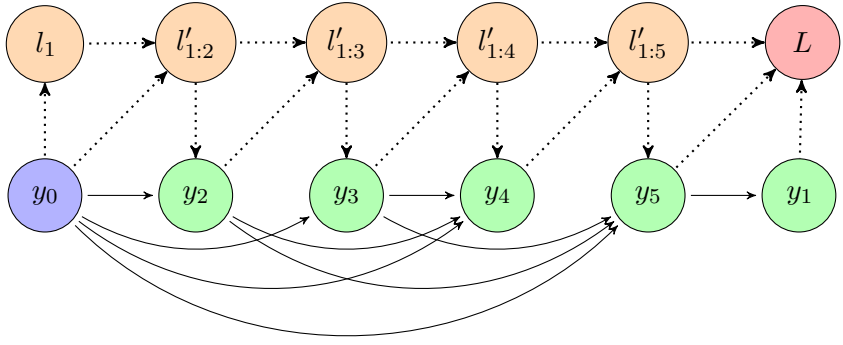

We select the first subset that has the highest aggregate score within $\{h_1,\dots, h_m\}$. Assume it is $\tilde{S}_j^{(t+1)}$ with score $h_j$. We then delete ${T_j:h_j}$ from B, add $T_j$ to A, and add the top $n$ lines from $\tilde{S}_j^{(t+1)}$ to ${S}^{(t+1)}$. Then, for each iteration $ > 1$, we calculate a set of temporary new scores  $\tilde{B}=\{T_1:\tilde{h_1},\dots,T_i: \tilde{h_i}, \dots T_m:\tilde{h_m}\}$ where each $\tilde{h_i}$ is the original score $h_i$ multiplied by   $\sum_{T_k \in A} \|T_i-T_k\|_{div}$, which is the sum of the diversity scores between $T_i$ and all previously chosen templates in $A$.  These scores are designed to strike a balance between finding high probability lines (as approximated by $h$) and lines whose templates have high diversity from the previously chosen templates (as measured by $\sum_{T_k \in A} \|T_i-T_k\|_{div}$). Afterwards, we repeat the process of choosing the template with the highest $\tilde{h}$ score, delete it from $B$,  add it to $A$, and add the top $n$ lines from its corresponding subset to ${S}^{(t+1)}$. We stop the iteration before the size of $S^{(t+1)}$ exceeds $N$.

Empirically, MTBS does not favor the templates with the rarest POS (largest distance from the rest), since those will have very low scores from $H(\cdot)$. It turns out MTBS picks templates that are reasonably different from each other while ensuring their generated lines have enough high scores. 


\subsection{Storyline Algorithm}\label{ssec:method-storyline}

We define the storyline of a limerick to be the last words of each line, denoted as $Y = (y_1, y_2, y_3, y_4, y_5)$, where $y_1$ is traditionally a name or place. In addition to making sure Y has an ``AABBA'' rhyming pattern, our storyline algorithm also helps \textit{LimGen} to maintain a consistent theme throughout its process of limerick generation. 

We define the probabilistic distribution of storyline $Y$ given a prompt word $y_0$ as: 
\begin{equation}\label{eq:prob_dist}\begin{split}
p(Y | y_0)
&=
p(y_2 | y_0) p(y_3 | y_0) p(y_4 | y_0, y_2, y_3)
\\&\cdot
 p(y_5 | y_0, y_2, y_3)p(y_1 | y_5)
,
\end{split}\end{equation}
\vspace{-0.2in}
\begin{align}
p(y_2 | y_0)
&\propto
\mathsf{Sim}(y_2, y_0)
,\nonumber\\
p(y_3 | y_0)
&\propto
\mathsf{Sim}(y_3, y_0)
,\nonumber\\
p(y_4 | y_0, y_2, y_3)
&\propto
\one_{y_4, y_3}^{(r)} \sum_{i \in \{0, 2, 3\}} \mathsf{Sim}(y_4, y_i)
,\nonumber\\
p(y_5 | y_0, y_2, y_3)
&\propto
\one_{y_5, y_2}^{(r)} \sum_{i \in \{0, 2, 3\}} \mathsf{Sim}(y_5, y_i)
,\nonumber\\
p(y_1 | y_5)
&\propto
\one_{y_1, y_5}^{(r)} \cdot \one_{y_1}^{(p)}
.\label{eq:cond_prob}
\end{align}
where the conditional distribution of each storyline word $y_i$ is a multinomial distribution over $\WW$. 
$\mathsf{Sim}(w_1, w_2)$ calculates the semantic similarity between words $w_1, w_2 \in \WW$, which is their distance in a pretrained word embedding space. Indicator function $\one_{w_1, w_2}^{(r)}$ denotes whether $w_1$ rhymes with $w_2$ and $\one_{w_1}^{(p)}$ denotes whether $w_1$ is a person's name. By sampling the storyline from $p(Y| y_0)$, we guarantee the following:\\
    - $y_2$ and $y_3$ are semantically related to $y_0$;\\
    - $y_4$ rhymes with $y_3$; $y_5$, $y_1$ and $y_2$ rhyme;\\
    - $y_4, y_5$ are semantically related to $y_0, y_2,y_3$.\\
Examples of samples from Storyline's distribution are provided in Table \ref{tab:eg_storyline}.

\begin{table}[ht]
\centering\small
\begin{tabular}{l|l}
\hline
Prompt & Storyline \\
\hline
war & (Wade, raid, campaign, again, stayed) \\
sports & (Pete, street, school, pool, athlete) \\
monster & (Skye, guy, scary, carry, pie) \\
forest & (Shea, day, friend, end, way) \\
\hline
\end{tabular}
\caption{Examples of storyline samples.}
\label{tab:eg_storyline}
\end{table}

During the process of generating a limerick, the Storyline Algorithm will sequentially generate many storylines in the order of $y_2, y_3, y_4, y_5, y_1$, each of which satisfies not only the rhyming constraint but also the constraints on POS template, syllable count and anapestic meter pattern. Figure \ref{fig:storyline_graph} shows the directed acyclic graph for the Storyline algorithm when the beam size of MTBS is 1.

In general, given a prompt word $y_0$, we start by generating the first line $l_1$, which has a canonical form, with a random name filled at its end as a placeholder for $y_1$ (otherwise, a pre-specified name will limit the options for $y_2, y_5$).
Following $l_1$, we use \textit{LimGen} to generate a set of second lines $\{\dots, l_2, \dots \}$ (as described in Algorithm \ref{algo:LimGen}) with last word left blank. We use $l_{1:2}'$ to denote a limerick generated up to this point (i.e., $l_1$ concatenated with an almost-finished $l_2$). For each $l_{1:2}'$, we repeatedly sample $y_2$ from the conditional Storyline distribution $p(y_2 | y_0)$ in \eqref{eq:cond_prob} until it satisfies constraints on POS, syllable and meter. $[l_{1:2}', y_2]$ together form the complete first two lines of a limerick. Continuing to generate lines, we use MTBS (described in Algorithm \ref{algo:MTBS}) to maintain a set of high-scoring and diverse second lines $\{\dots,[l_{1:2}',y_2],\dots\}$. Note that our language model also assigns a probability score for each $y_2$. We can continue generating $l_{1:k}'$ with \textit{LimGen} and sampling $y_k$ from the conditional Storyline distribution for $k = 3, 4, 5$ in a similar fashion.
Finally, we sample $y_1$ from $p(y_1 | y_5)$ and replace the random name at the end of $l_1$ by it. The result is a set of limericks $\{\dots,L,\dots \}$, from which we choose the highest scoring one.

\section{Experiment}\label{sec:exp}

\subsection{Experimental Setup}
To implement \textit{LimGen}, a significant amount of effort has gone into adapting existing NLP technologies for limerick generation. In order to extract POS templates from human written limericks, we modified the POS categories in NLTK \citep{nltk} by refining certain categories for better quality in our generated limericks. Leveraging NLTK's  POS tagging technique, we obtained a list of refined POS templates from a small limerick dataset of 200 human-written limericks from \citet{brownie, familyfriend}. For a full list of our modified POS categories see \citet{ourcode}. Since the filtering function $\mathcal{F}$ requires knowing each word's syllable and stress pattern, we use \citet{cmu} for information on syllable and stress.

As for the implementation of the Storyline algorithm, there are several existing technologies to indicate whether two words rhyme with each other. For example, \textit{Deep-speare} \cite{lau-etal-2018-deep} proposed a character-level LSTM to learn rhyming. For the sake of simplicity and accuracy, we used a rhyming dictionary curated from \citet{datamuse}. We also used a dictionary of names \citep{rhyming-name} from which the Storyline algorithm can choose $y_1$, the name in the poem's first line. To calculate the semantic similarity between two words, we use the pre-trained word embedding space from \textit{spaCy}'s model \citep{spacy2}.

Note that Algorithm \ref{algo:LimGen} is only responsible for generating the last four lines of a limerick. Since first lines of limerick usually have a canonical form, we generate the first lines separately. 

The outline of this section is as follows. We first show why GPT-2 -- or even retrained GPT-2 -- cannot produce limericks without \textit{LimGen}. We then show the low-quality output from prior attempts at limerick generation. We have also designed five experiments to compare the quality of \textit{LimGen}'s output with limericks from human poets, baseline algorithms and other state-of-the-art poetry systems re-purposed for limerick generation. All five experiments were evaluated on Amazon Mechanical Turk by crowd-workers, following a protocol similar to that of \citet{lau-etal-2018-deep,hopkins2017automatically} (see Section \ref{ssec:exp-single} for details).
Additionally, an ``Expert Judgement'' experiment was conducted where more experienced judges directly evaluated the performance of \textit{LimGen}'s output and human-written limericks across a variety of metrics (See Section \ref{ssec:exp5} for details). 

Since \textit{LimGen} has three major components: AMTC, MTBS and Storyline, we designed three baseline algorithms for an ablation analysis in order to investigate the effectiveness of each of them.\\
    -\textit{Single-Template}: MTBS+Storyline but without AMTC\\
    -\textit{No-Story}:
    AMTC+MTBS but without pre-selected storylines\\
    -\textit{Candidate-Rank}:
    AMTC+Storyline but we have replaced the MTBS algorithm with another modified beam search algorithm Candidate-Rank \cite{degradation-cohen19}

In our experiments, \textit{LimGen} and all baseline algorithms use a total beam size of $N=360$ at each step, MTBS algorithm's individual beam size per template is $n=12$, and we take the highest scoring poem from the set of output poems. For implementation details please refer to our online GitHub repository \cite{ourcode}. 

\subsection{GPT-2 cannot generate poems by itself}
\label{ssec:naive-gpt}

\begin{table}[ht]
\centering\footnotesize
\begin{subtable}{0.49\textwidth}
\centering
\caption{Output of na\"ive GPT-2 generation}
\begin{tabular}{l}
\hline
There was a kind girl whose name is Jane,\\
A girl who I did not know,\\
He then added,\\
She had tons of luggage,\\
It seemed I could walk where she.\\
\hline
\end{tabular}
\vspace*{5pt}
\caption{This output is an exact replica of a human limerick \citep{diverse-limericks} in the training corpus of GPT-2.}
\begin{tabular}{l}
\hline
Wait, there was a young lady in china,\\
Who was quite a greedy young diner.\\
She feasted on snails,\\
Slugs, peacocks and quails,\\
`No mixture,' she said, 'could be finer.' \\
\hline
\end{tabular}
\end{subtable}
\caption{Two examples of Na\"ive GPT-2.}
\label{naive_examples}
\end{table}

A na\"ive implementation of GPT-2 simply cannot produce original and valid limericks. GPT-2 tends to generate long sentences that exceed the syllable limit for limericks. To meet a syllable constraint, we would need to truncate the generated sentences, which creates lines that do not end correctly. Rhyming is insurmountable if we do not utilize additional algorithms, as evidenced by Example (a) of Table \ref{naive_examples}. The output lacks poetic quality since the training corpus of GPT-2 does not mainly consist of limericks or other kinds of poetry. 

If we try to re-train the last few layers of a GPT-2 model on our entire limerick dataset, it does not solve the problem. To our knowledge, our entire dataset is the biggest and most comprehensive limerick dataset, consisting of more than 2000 limericks from several sources \cite{brownie,familyfriend, lear-limericks, penguin-limericks, australia-limericks}. Even though this dataset is much larger than the subset of data  ($\approx 300$) from which we extracted templates, it is still insufficient to retrain GPT-2. The result of re-training is that GPT-2 severely overfits. It only regurgitates limericks from the training corpus, as seen in Example (b) of Table~\ref{naive_examples}. 

Terminating training early (in order to avoid memorization or overfitting) leads only to an awkward merge of problems shown in the two examples of Figure~\ref{naive_examples} in which the model has not learned enough to faithfully reproduce the form of a limerick, but also often loses coherence abruptly or regurgitates the training set. 

Just as \textit{LimGen} needs a powerful pre-trained language model such as GPT-2, without \textit{LimGen}'s algorithms, GPT-2 by itself is unable to accommodate the constraints of limerick generation due to the deficiency of training data.
\begin{figure*}[t]
    \centering
    \includegraphics[width=\textwidth]{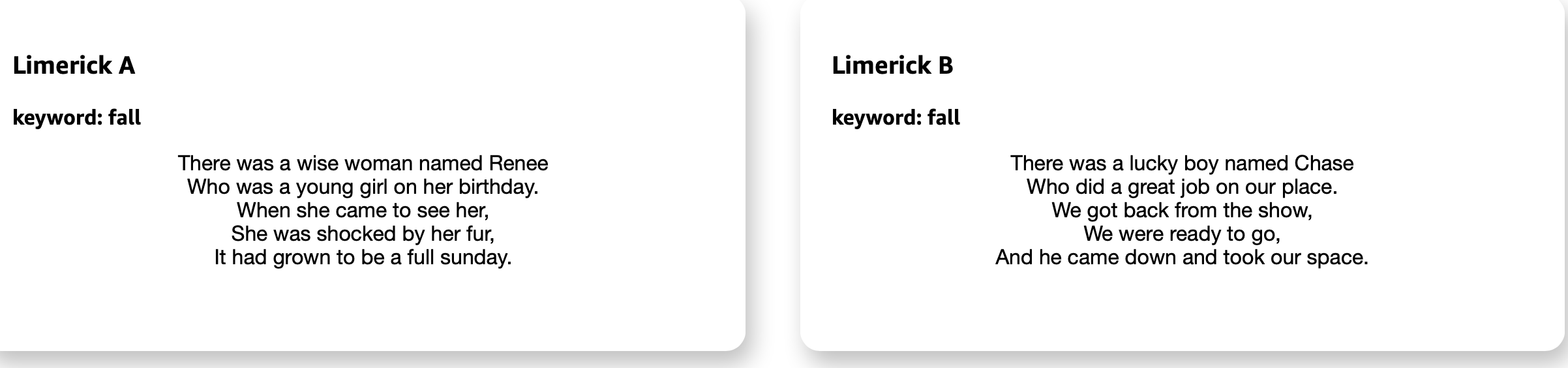}
    \caption{Side-by-side comparison of two limericks generated from different methods}
    \label{fig:ABlimericks}
\end{figure*}

\subsection{Prior attempts at Limerick Generation}
\label{ssec:prior-attempt}
\begin{table}[ht]
\centering\footnotesize
\begin{subtable}{0.49\textwidth}
\centering
\caption{Example of \cite{levy2001computational}'s system}
\begin{tabular}{l}
\hline
Ears christmas new throat boat apparel,\\
Plain always obsessed deal idea,\\
Attempt blast work many,\\
Mercator aghast,\\
Kathleen mind revealed barge bugs humor.\\
\hline
\end{tabular}
\end{subtable}
\begin{subtable}{0.49\textwidth}
\centering
\caption{Example of \cite{manurung}'s system}
\begin{tabular}{l}
\hline
The bottle was loved by Luke.\\
a bottle was loved by a dog\\
A warm distinctive season humble mellow,\\
smiled refreshingly slowly. Ran.\\
\hline
\end{tabular}
\end{subtable}
\caption{Two prior attempts at limerick generation}
\label{prior_examples}
\end{table}
\begin{table}[ht]
\centering\footnotesize
\begin{subtable}{0.49\textwidth}
\centering
\caption{Example of \cite{poem-generator}}
\begin{tabular}{l}
\underline{There once was a} \textbf{man} \underline{called} \textbf{Liam}.\\
\underline{He said, "See the} coliseum!",\\
\underline{It was rather} young,\\
\underline{But not very} zedong,\\
\underline{He couldn't resist the} mit im.\\
\end{tabular}
\end{subtable}
\begin{subtable}{0.49\textwidth}
\centering
\caption{Example of \cite{poem-of-quotes}}
\begin{tabular}{l}
\underline{There was a} \textbf{man} \underline{from} White\\
\underline{Who liked to} \textbf{fly his kite}\\
\underline{On each sunny day}\\
\underline{The man would say}\\
\underline{'Oh, how I miss} White!'\\
\end{tabular}
\end{subtable}
\caption{Examples of internet poem generators. Underlined parts are human-written half sentences and bold parts are user inputs.}
\label{internet_examples}
\end{table}
\citet{levy2001computational} stated that ``the current system produces limericks that in many ways seem random.'' We have re-run their implementation, and it only produced meaningless verses with serious grammatical issues. \citet{manurung} stated that their work is unfinished and stated that their results ``can hardly be called poems'' (see examples in Table \ref{prior_examples}). Empirically, \textit{LimGen} has a clear advantage over both prior works. Therefore, the low-quality output from these system do not warrant an extensive comparison with \textit{LimGen}'s poems. 

On the other hand, popular internet poem generators \cite{poem-generator,poem-of-quotes} have a set of human-written half-finished sentences that are assembled with user input words to create limericks (see Table \ref{internet_examples}). However, because so little of the resulting limerick is generated by a machine, we cannot consider these internet poem generators as automatic limerick generation systems. 


\subsection{Experiment 1: \textit{LimGen} vs$.$ \textit{No-Story}}\label{ssec:exp-single}
As we have mentioned before, the \textit{No-Story} baseline still utilizes the AMTC and MTBS algorithms. This experiment demonstrates the importance of having pre-selected storylines in poetry generation. 

\begin{figure}[ht]
\centering
\includegraphics[width=0.49\textwidth]{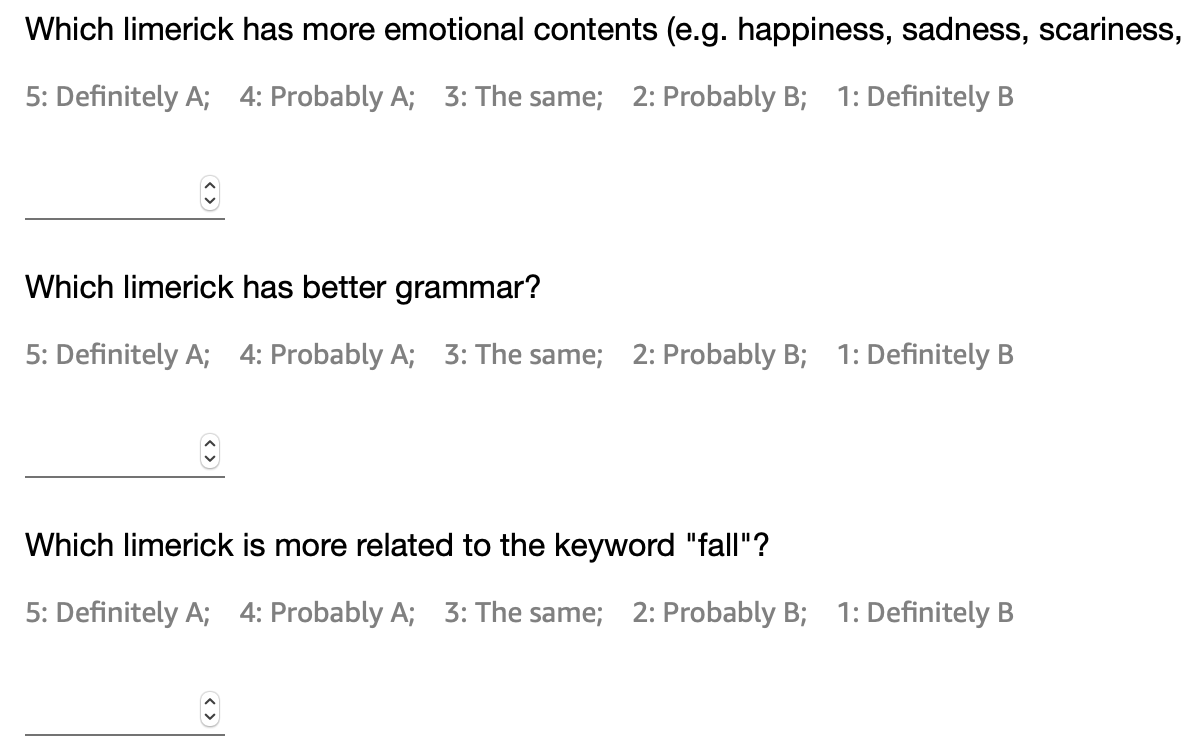}
\includegraphics[width=0.49\textwidth]{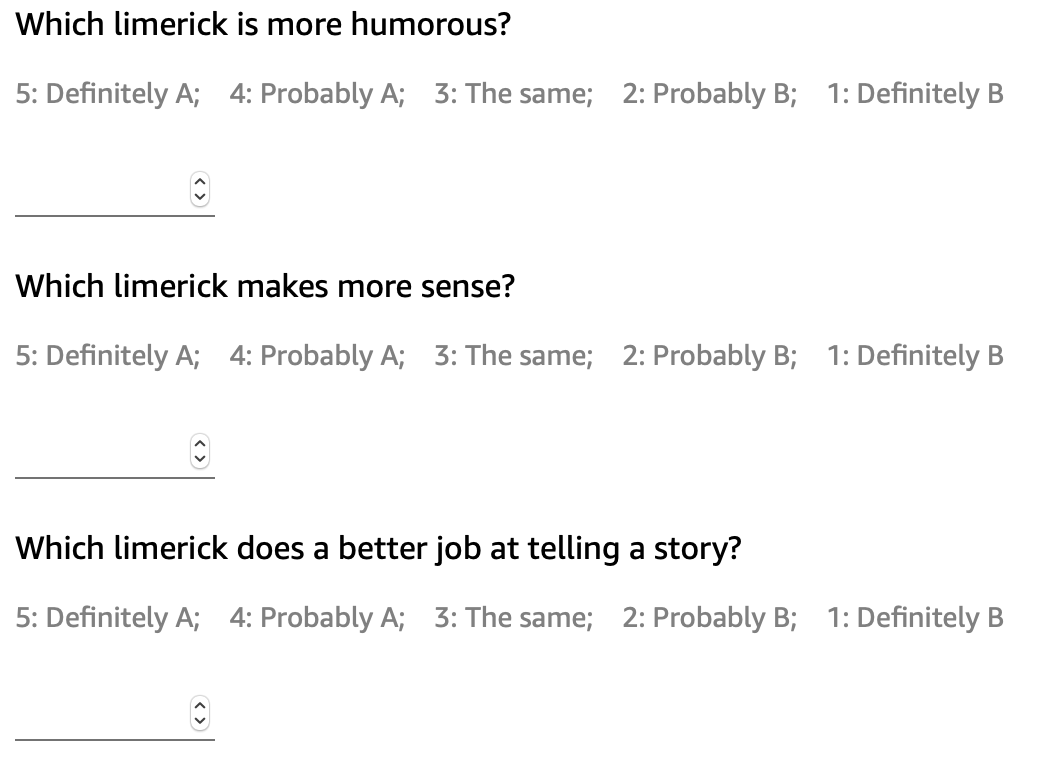}
\caption{The list of questions on Amazon Mechanical Turk survey}
\label{fig:qesutions12}
\end{figure}

We randomly selected 50 pairs of limericks, in which each pair of limericks consists of one generated by \textit{LimGen} and another generated by \textit{No-Story} using the same prompt word. For each pair of limericks, 5 different crowd-workers (each with an approval rate $\geq 90\%$) answered a list of 6 questions on different evaluation metrics (humor, sensibleness, story-telling, emotional content, grammar, thematic relatedness to prompt) and an additional sanity-check question to filter out illogical responses. Figure \ref{fig:ABlimericks} and Figure \ref{fig:qesutions12} shows the side-by-side comparison of a pair of limericks and the list of questions exactly as they appeared on the Amazon Mechanical Turk survey. Note that the output of \textit{LimGen} has a 50\% chance of being either Limerick A or B to avoid any left-right bias.

A total of 250 response were recorded, and a small number of responses were filtered out since they did not answer the sanity check question correctly, which asks crowd-workers to count the number of 3-letter words in the fourth line of Limerick B. We have transformed the response such that a response of 5 always means that the poem is rated as ``Definitely \textit{LimGen}'s output;'' i.e., if \textit{LimGen} produced Limerick B, we transform 5 to 1, 4 to 2, 2 to 4 and 1 to 5. After this transformation, we calculated the mean and standard deviation for each metric. Since all questions ask crowd-workers to compare the qualities of two limericks, the results are relative. It should be clear that for any metric, an average greater 3 means \textit{LimGen} is performing better than the baseline method on that metric. To be precise, if the mean of a metric is $>3$, we run a one-sided t-test with the null-hypothesis being ``metric $\leq 3$, i.e., \textit{LimGen} is not doing better than baseline.'' If the mean of a metric is $<3$, suggesting the baseline is probably doing better, we run the complementary one-sided t-test with the null-hypothesis being ``metric $\geq 3$, i.e., baseline is not doing better than \textit{LimGen}.''

\begin{table}[ht]
\centering
\tabcolsep=0.11cm
\begin{tabular}{|l|c|c|c|}
\hline
\backslashbox{Metrics}{Statistics} & mean & sd & p-value \\
\hline
emotion & 3.03  & 1.22 & 0.38 \\
\hline
grammar & \textbf{3.18}  & 1.27 & 0.03 \\
\hline
humor & \textbf{3.14}  & 1.20 & 0.05 \\
\hline
relatedness & \textbf{3.32}  & 1.22 & 2.0$\times10^{-4}$ \\
\hline
story-telling & \textbf{3.35}  & 1.38 & 3.0$\times10^{-4}$ \\
\hline
sensibleness & 3.14  & 1.42 & 0.09 \\
\hline
\end{tabular}
\caption{\textit{LimGen} vs$.$ \textit{No-Story}}
\label{tab:exp1}
\end{table}

From Table \ref{tab:exp1}, the p-value of grammar, humor, relatedness to prompt, and story-telling are all small enough to reject the null hypothesis, which shows that \textit{LimGen} was better than \textit{No-Story} in all four categories. We can  weakly reject the null hypothesis for the sensibleness metric, which shows that \textit{LimGen} also may outperform \textit{No-Story} with respect to sensibleness.  However, the p-value of emotion is 0.38, therefore we cannot claim \textit{LimGen}'s output has better emotional content than \textit{No-Story}. Overall, we see that \textit{LimGen} empirically outperforms \textit{No-Story} in 5 categories. From this experiment, we see that having pre-selected storylines not only makes the limerick more related to the prompt (as expected), but it also enhances the consistency of story-telling and other important poetic metrics. 

All other experiments were designed in the same way as Experiment 1.

\subsection{Experiment 2: \textit{LimGen} vs$.$ \textit{Single-Template}}
\label{ssec:exp2}
As we have mentioned before, the \textit{Single-Template} baseline still utilizes the MTBS and Storyline algorithms. However, we have designed the \textit{Single-Template} baseline so that it mimics a traditional rule-based poetry generation algorithm, wherein a single POS template is followed \cite{colton2012}. For each prompt word, a random template is selected and \textit{Single-Template}  generates text according to it. This experiment will highlight the advantages of adaptively choosing templates. 

\begin{table}[ht]
\centering
\tabcolsep=0.11cm
\begin{tabular}{|l|c|c|c|}
\hline
\backslashbox{Metrics}{Statistics} & mean & sd & p-value \\
\hline
emotion & \textbf{3.20}  & 1.23 & 0.02 \\
\hline
grammar & \textbf{3.48}  & 1.28 & 4.4$\times 10^{-6}$ \\
\hline
humor & \textbf{3.27}  & 1.16 & 0.001 \\
\hline
relatedness & 2.95  & 1.42 & 0.34 \\
\hline
story-telling & \textbf{3.52}  & 1.39 & 1.3$\times10^{-6}$ \\
\hline
sensibleness & \textbf{3.40}  & 1.36 & 9.8$\times10^{-5}$ \\
\hline
\end{tabular}
\caption{ \textit{LimGen} vs$.$ \textit{Single-Template}}
\label{tab:exp2}
\end{table}

From Table \ref{tab:exp2}, we see that the means of 5 metrics are significantly greater than 3, which means AMTC has a clear advantage over using a single template constraint. This makes sense, since AMTC allows \textit{LimGen} to adaptively choose which template to follow.
Though AMTC is easy to implement, we see substantial improvement over its predecessors. Lastly, the mean of relatedness is 2.95, but the p-value is not small enough to claim that \textit{LimGen} is worse than \textit{Single-Template}.

\subsection{Experiment 3: \textit{LimGen} vs$.$ \textit{Candidate-Rank}}
\label{ssec:exp3}
Candidate-Rank beam search  \citep{degradation-cohen19} addressed the degradation of beam search performance when the beam size grows too large. It is simple to implement, and remains one of the best modified beam search algorithms. 

\begin{figure}[ht]
\centering
\includegraphics[width=\linewidth]{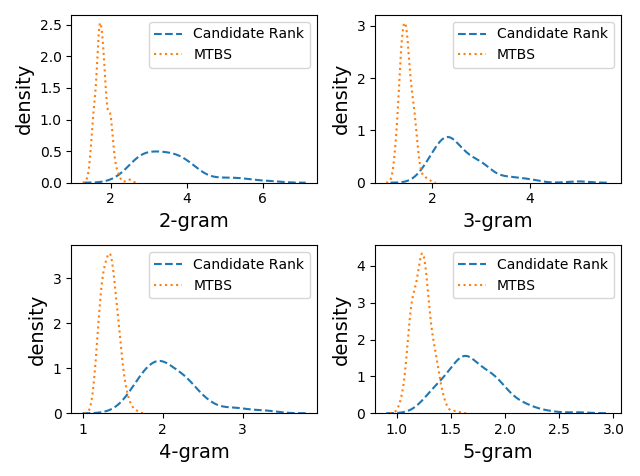}
\caption{Distributions of ``mean popularity of $n$-gram'' within last lines for \textit{LimGen} and \textit{Candidate-Rank} output.}
\label{fig:n-gram}
\end{figure}

\begin{table}[ht]
\centering
\tabcolsep=0.11cm
\begin{tabular}{|l|c|c|c|}
\hline
\backslashbox{Metrics}{Statistics} & mean & sd & p-value \\
\hline
emotion & 2.88  & 1.20 & 0.62 \\
\hline
grammar & 3.06  & 1.14 & 0.25 \\
\hline
humor & 2.91  & 1.15 & 0.15 \\
\hline
relatedness & 3.03  & 1.22 & 0.37 \\
\hline
story-telling & 3.06  & 1.31 & 0.28 \\
\hline
sensibleness & \textbf{3.19}  & 1.27 & 0.034 \\
\hline
\end{tabular}
\caption{ \textit{LimGen} vs$.$ \textit{Candidate-Rank}}
\label{tab:exp3}
\end{table}

From Table \ref{tab:exp3}, the only statistically significant result is that \textit{LimGen} outperforms \textit{Candidate-Rank} with respect to sensibleness, which is due to the diversity fostering beam search MTBS. Since in our left-to-right limerick generation procedure, \textit{LimGen} picks the next word that not only satisfies POS, meter, syllable and rhyming constraints but also flows naturally with the preceding lines, it is beneficial to maintain a diverse set of preceding partial lines to choose from. This ensures coherency and sensibleness in the output limericks.
We can see the role of MTBS in fostering diversity more explicitly by counting distinct POS templates and by calculating the repetition (in terms of $n$-grams) within a fixed number of output limericks. For both \textit{LimGen} and \textit{Candidate-Rank}, a maximum of 360 sentences can be processed in parallel. We ran both methods 200 times (using 100 prompt words, each with one female and one male name). \textit{LimGen} has an average of 27 different templates per $\sim$200 poem run, whereas \textit{Candidate-Rank} only used 6 templates on average. 
For each run, to measure diversity, we randomly selected 50 limericks from the output set and calculated the ``mean popularity of each $n$-gram'' (e.g., 2-gram, 3-gram, 4-gram, 5-gram) in their last lines.
Specifically, for each n-gram ($n$ consecutive words) within those 50 last lines, we record its number of occurrences within those 50 lines. We then average all those recorded numbers and denote it as the ``mean popularity of $n$-gram.'' For instance, ``mean popularity of 3-gram''$=2.0$ indicates that, on average, each 3-gram within those 50 lines repeats twice. A high value of the ``mean popularity of $n$-gram'' indicates heavy phrase repetition. As we can see from Figure \ref{fig:n-gram}, MTBS has a significantly lower ``mean popularity of $n$-gram'' than the Candidate-Rank beam search, which indicates more sentence diversity within MTBS' output. 


\subsection{Experiment 4: \textit{LimGen} vs$.$ \textit{Deep-speare}}
\label{ssec:exp4}
Similar to GPT-2, \textit{Deep-speare}'s language model was trained on 2000 limericks for 30 epochs until validation loss stopped decreasing using the optimal hyper-parameters provided by \citet{lau-etal-2018-deep}. Since the pentameter model for stress and the rhyming model in the full \textit{Deep-speare} are not guaranteed to adhere to limericks' stress, syllable and rhyming constraints, especially when the training data are scarce, we replaced these two models (pentameter and rhyming) with  constraints to ensure the output from \textit{Deep-speare} meets the requirements of limericks. Compared to GPT-2, \textit{Deep-speare} is a much smaller model. In the original paper, it was trained on only 7000 quatrains of sonnets. After training on our limerick dataset, it was able to produce some form of limerick that warrants a comparative experiment. 
\begin{table}[ht]
\centering
\tabcolsep=0.11cm
\begin{tabular}{|l|c|c|c|}
\hline
\backslashbox{Metrics}{Statistics} & mean & sd & p-value \\
\hline
emotion & \textbf{3.46}  & 1.18 & 2.96$\times 10^{-7}$ \\
\hline
grammar & \textbf{4.02}  & 1.35 & $\approx 0$ \\
\hline
humor & \textbf{3.36}  & 1.24 &  6.74$\times 10^{-5}$\\
\hline
story-telling & \textbf{3.98}  & 1.11 & $\approx 0$ \\
\hline
sensibleness & \textbf{3.99}  & 1.18 & $\approx 0$ \\
\hline
\end{tabular}
\caption{ \textit{LimGen} vs$.$ \textit{Deep-speare}}
\label{tab:exp4}
\end{table}

We can clearly see from Table \ref{tab:exp4} that for the task of limerick generation, \textit{LimGen} outperforms this adapted version of \textit{Deep-speare} (which is considered a state-of-the-art neural network for English poetry generation) across all metrics. It remains to be seen whether \textit{Deep-speare} will improve given more training data. However, it is unclear where more data would come from.

\subsection{Experiment 5: \textit{LimGen} vs$.$ Human Poets}
\label{ssec:exp5}
\begin{table*}[ht]
\centering\footnotesize
\begin{subtable}{0.49\textwidth}
\centering
\begin{tabular}{l}
\hline
There once was a brave soldier named Wade\\
Who led a small army on his raid.\\
He died on the campaign,\\
His body burned again,\\
But he kept his promises and stayed.\\
\hline
\end{tabular}
\caption{Prompt word: war}
\end{subtable}
\begin{subtable}{0.49\textwidth}
\centering
\begin{tabular}{l}
\hline
There was a honest man named Dwight\\
Who lost all his money in a fight.\\
His friends were so upset,\\
They were willing to bet,\\
And they did not like feeling of spite.\\
\hline
\end{tabular}
\caption{Prompt word: loss}
\end{subtable}

\begin{subtable}{0.49\textwidth}
\centering
\begin{tabular}{l}
\hline
There was a loud waitress named Jacque,\\
Who poured all her coffee in a shake.\\
But the moment she stirred,\\
She was struck by a bird,\\
Then she saw it fly towards the lake.\\
\hline
\end{tabular}
\caption{Prompt word: shaken}
\end{subtable}
\begin{subtable}{0.49\textwidth}
\centering
\begin{tabular}{l}
\hline
There once was a nice man named Theodore\\
Who killed all his family in a war.\\
He came back from the dead,\\
With a scar on his head,\\
But he lost his memories and more.\\
\hline
\end{tabular}
\caption{Prompt word: violent}
\end{subtable}
\caption{More Example limericks from \textit{LimGen}}
\label{tab:eg_exp}
\end{table*}
In this experiment, 50 human limericks were chosen randomly from our database. Although not completely homogeneous in their poetic qualities, they were all well-thought-out and well-written, and represent genuine effort from their authors.   

\begin{table}[ht]
\centering
\tabcolsep=0.11cm
\begin{tabular}{|l|c|c|c|c|}
\hline
\backslashbox{Metrics}{Statistics} & mean & sd & p-value & $> 3$\\
\hline
emotion & 2.84  & 1.41 & 0.04 & 43\%\\
\hline
grammar & 2.97  & 1.41 & 0.29 & 58\%\\
\hline
humor & 2.21  & 1.41 &  $\approx 0$ & 22\%\\
\hline
story-telling & 2.55  & 1.49 & $\approx 0$ & 37\%\\
\hline
sensibleness & 2.58  & 1.47 & $\approx 0$& 35\%\\
\hline
\end{tabular}
\caption{ \textit{LimGen} vs$.$ Human Poets}
\label{tab:exp5}
\end{table}

In Table \ref{tab:exp5}, we added a column that records the percentage of limerick pairs with an average response $> 3$, i.e., the percentage of \textit{LimGen}'s limericks that are better than human's on a specific metric according to crowd-workers. Clearly, human poets outperform \textit{LimGen} on several metrics. It is not statistically conclusive which method is better with respect to grammar, presumably due to the template-guided approach that ensures grammatical correctness. Upon careful inspection, we noticed that for several metrics, there is actually a significant portion of \textit{LimGen}'s output that were rated more highly than human-written limericks. For example, 43\% of the machine-generated limericks had better emotional content than human poems. Another observation is that humor seems to be the hardest attribute for \textit{LimGen} to emulate and master. Even though \textit{LimGen} does output humorous limericks at times, they usually do not have the highest score according to our scoring function $H(\cdot)$; in other words, even though humorous poems were generated, our scoring mechanism could not recognize them as humorous. 

\begin{figure}[ht]
\centering
\includegraphics[width=\linewidth]{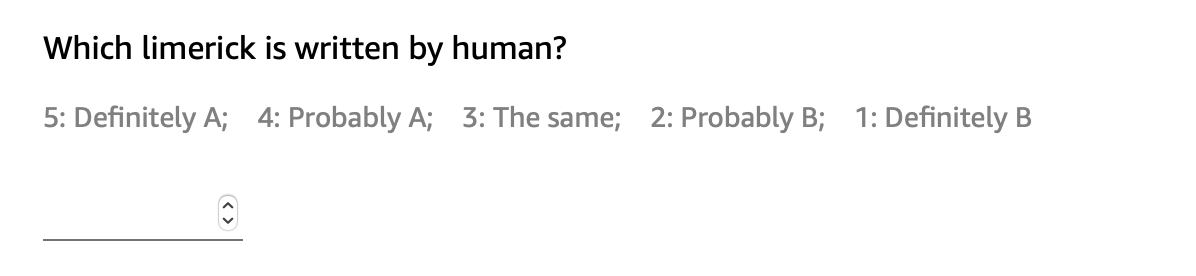}
\caption{The Turing test question}
\label{fig:turing-question}
\end{figure}

In this same experiment, we asked crowd-workers a Turing test question for each limerick pair (one by a human and one by \textit{LimGen}) (Figure \ref{fig:turing-question}): whether Limerick A or B is more likely to be written by a human.  Recall that in our analysis we have transformed the data such that a score of 1 indicates the crowd-worker thinks that the poem was surely  written by machine. The recorded score distribution is $1:11\%, 2:14\%, 3:18\%, 4:29\%,5:27\%$. 
Scores 1 and 2 are when \textit{LimGen}'s limericks are mistaken as human-written when directly compared with actual human-written poems. Score 3 is when judges cannot differentiate between \textit{LimGen}'s output and human poems. \textit{Overall, the crowd-workers cannot differentiate LimGen's output from human-written poems 43\% of the time. }

While so far we have compared \textit{LimGen} with baselines and prior works on a relative scale because people are better at comparing items rather than assigning direct values to them, we now evaluate \textit{LimGen}'s output on an absolute scale, which would paint a clearer picture of its strength and weakness on poetic metrics. We convened an expert panel of 20 Duke students who are proficient in English, have received a liberal arts education and have completed two college-level courses designated to satisfy the literature requirement of the university. Since the intended audience of limericks is the general public, we believe that these panelists, with their considerable experience and expertise in the English language,  are qualified to directly evaluate 60 limericks (30 from \textit{LimGen} and 30 from humans) across the same metrics on an absolute scale from 1 to 5 (1 being the worst and 5 being the best). Each panelist completed at least one assignment, which consists of 6 poems randomly chosen from the set of 60 limericks. We ensured that each limerick was evaluated at least twice and the panelists did not see repeated limericks. None of these panelists knew anything about how the automated poems were generated. They were only notified that they would see a mixture of machine and human-written limericks. 

\begin{table}[ht]
\centering
\tabcolsep=0.11cm
\begin{tabular}{|l|c|c|c|}
\hline
& Human & \textit{LimGen} & p-value\\
\hline
emotion & 3.79 $\pm$ 0.98 & 2.93$\pm$ 0.99& 0.006\\
\hline
grammar & 4.22  $\pm$ 0.99 & 3.65 $\pm$ 0.96 & 0.068\\
\hline
humor & 3.92 $\pm$ 1.01 & 2.21 $\pm$ 0.92 & $\approx$ 0\\
\hline
story-telling & 4.44 $\pm$ 0.74 & 3.68 $\pm$ 0.85 & 0.009\\
\hline
sensibleness & 3.88 $\pm$ 0.92 & 3.03 $\pm$ 1.05& 0.006\\
\hline
\end{tabular}
\caption{Expert Judges: \textit{LimGen} vs$.$ Humans}
\label{tab:exp6}
\end{table}
The scores in this survey are absolute values rather than relative values. We interpret an average over 3 on a metric as a decent level of performance. From Table \ref{tab:exp6}, although expert judgement confirms that human poets outperform \textit{LimGen}, it still shows that \textit{LimGen} performs decently according to several metrics: \textit{LimGen} has decent grammar and can tell a story well with its verses. It seems that grammar and story-telling are the easiest poetic attributes to master, since both human poets and \textit{LimGen} have the highest scores on these metrics. Emotion and sensibleness are harder to learn. But what really differentiates human poets and \textit{LimGen} is poets' ability to consistently make jokes. 

Overall, we find our results encouraging, as they not only show that \textit{LimGen} outperforms all prior baselines by a clear margin, but also shows that \textit{LimGen} has the potential to approach human level performance in the future. More outputs from  \textit{LimGen} are in Table \ref{tab:eg_exp}.


\section{Conclusion}\label{sec:con}
\textit{LimGen} is the first fully-automated limerick generation system. Using human judgements, we have shown that our adaptive multi-templated constraints provide \textit{LimGen} with a combination of quality and flexibility.  We have shown the value of our diversity-fostering multi-templated beam search, as well as the benefits of our Storyline algorithm. 





\section{Acknowledgements}

We would like to extend our sincere appreciation to all people involved in this research project, especially our colleagues Matias Benitez, Dinesh Palanisamy and Peter Hasse for their support and feedback in the initial stage of our research. We would also like to thank Alstadt for funding. We have included a few more poems from \textit{LimGen} in Table \ref{supplement-examples}. Please refer to our online GitHub repository \citep{ourcode} for implementation details and more poems. 

\begin{table}[ht]
\centering\footnotesize
\begin{subtable}{0.49\textwidth}
\centering
\begin{tabular}{l}
\hline
There was a shy actor named Dario, \\
Who played a big role on our show. \\
He came back from the break, \\
And we went to the lake, \\
And he sat down and took his photo. \\
\hline
\end{tabular}
\caption{Prompt word: Season}
\end{subtable}
\begin{subtable}{0.49\textwidth}
\centering
\begin{tabular}{l}
\hline
There was a artist named Cole,\\
Who made a huge impact on my soul. \\
He was a musician, \\
He was on a mission, \\
And that is the beauty of this role.\\
\hline
\end{tabular}
\caption{Prompt word: Art}
\end{subtable}
\begin{subtable}{0.49\textwidth}
\centering
\begin{tabular}{l}
\hline
There once was a liar named Kai,\\
Who fooled a grand jury on her lie. \\
I had a suspicion, \\
I was on a mission, \\
I was ready to fight and to die.\\
\hline
\end{tabular}
\caption{Prompt word: Cunning}
\end{subtable}
\begin{subtable}{0.49\textwidth}
\centering
\begin{tabular}{l}
\hline
There was a bright cleaner named Dot,\\
Who put all her money in a pot. \\
When she started to smoke, \\
She was struck by a stroke, \\
She has a severe case of a clot.\\
\hline
\end{tabular}
\caption{Prompt word: Water}
\end{subtable}
\begin{subtable}{0.49\textwidth}
\centering
\begin{tabular}{l}
\hline
There was a funky chef named Dwight,\\
Who cooked a great meal on our night. \\
We got back from the bar, \\
And we walked to the car, \\
And we sat down and had our bite. \\
\hline
\end{tabular}
\caption{Prompt word: Beer}
\end{subtable}
\begin{subtable}{0.49\textwidth}
\centering
\begin{tabular}{l}
\hline
There was a cruel judge named Lyle,\\
Who killed a young girl on his trial. \\
It was like a nightmare, \\
I was scared by his stare, \\
But I knew his intentions and smile.\\
\hline
\end{tabular}
\caption{Prompt word: Death}
\end{subtable}
\caption{Additional limericks from \textit{LimGen}}
\label{supplement-examples}
\end{table}

\newpage
\bibliography{acl2020.bib}
\bibliographystyle{acl_natbib}





\end{document}